\documentclass[11pt]{article}
\usepackage{nodalida2025}
\usepackage{times}
\usepackage{hyperref}
\usepackage{times}
\usepackage{url}
\usepackage{latexsym}
\usepackage{graphicx}
\usepackage{todonotes}
\usepackage[linewidth=1pt]{mdframed}
\usepackage{lipsum}
\usepackage{subcaption}
\usepackage{tabularx}
\usepackage{listings}
\aclfinalcopy 

\title{Got Compute, but No Data: Lessons From Post-training a Finnish LLM}

\author{
  \textbf{Elaine Zosa\textsuperscript{1}}
  \textbf{Ville Komulainen\textsuperscript{2}}
  \textbf{Sampo Pyysalo\textsuperscript{2}}
\\
\\
  \textsuperscript{1}Silo AI, Finland
  \textsuperscript{2}TurkuNLP, University of Turku, Finland
\\
\tt{firstname.lastname@\{silo.ai,utu.fi\}}
}

\begin{document}
\maketitle
\begin{abstract}
As LLMs gain more popularity as chatbots and general assistants, methods have been developed to enable LLMs to follow instructions and align with human preferences. These methods have found success in the field, but their effectiveness has not been demonstrated outside of high-resource languages. In this work, we discuss our experiences in post-training an LLM for instruction-following for English and Finnish. We use a multilingual LLM to translate instruction and preference datasets from English to Finnish. We perform instruction tuning and preference optimization in English and Finnish and evaluate the instruction-following capabilities of the model in both languages. Our results show that with a few hundred Finnish instruction samples we can obtain competitive performance in Finnish instruction-following. We also found that although preference optimization in English offers some cross-lingual benefits, we obtain our best results by using preference data from both languages. We release our model, datasets, and recipes under open licenses at~\url{https://huggingface.co/LumiOpen/Poro-34B-chat-OpenAssistant}.
\end{abstract}

\section{Introduction}
Foundational LLMs are language completion models that need to be finetuned after pretraining to be able to respond to user questions and follow instructions~\cite{ouyang2022training}. This post-training process involves \textit{supervised finetuning} where the model is trained to act as an assistant by training on a dataset of prompt-response pairs. \textit{Preference optimization} further aligns the model to human preferences such as helpfulness, harmlessness, and honesty~\cite{bai2022constitutional}. These methods have resulted in LLMs becoming more capable of answering complex questions involving reasoning, coding, math, and science~\cite[e.g.,][]{dubey2024llama,jiang2024mixtral,team2024gemma}. The effectiveness of these methods, however, have not been demonstrated for smaller and less-resourced languages, such as Finnish.

One of the major challenges we face in post-training in smaller languages is the scarcity of training data. The situation is even more challenging for commercial settings as most of the datasets available today are generated by LLMs with restrictive licenses. The availability of evaluation benchmarks for chat models in small languages is also an issue. Popular benchmarks such as MTBench~\cite{zheng2024judging} and IFEval~\cite{zhou2023instruction} are designed for English models and have not been adapted for use in a multilingual setting.~\footnote{While revising this paper, a multilingual, multi-turn IFEval was released~\cite{he2024multi}, but it does not include Finnish.} 

In this paper, we discuss our experiences in post-training an LLM in Finnish and English. We use the LLM that we want to finetune to machine-translate instruction and preference datasets into Finnish. We use a commercial machine-translation service to translate a widely-used instruction-following evaluation (IFEval) benchmark into Finnish. We experimented with different combinations of Finnish and English data in instruction tuning and preference optimization. We also experimented with different methods to improve vanilla instruction tuning.

\section{Related Work}

The post-training of base LLMs, popularised in InstructGPT~\cite{ouyang2022training}, can be broadly divided into two categories: instruction tuning and preference optimization. Instruction tuning, also known as supervised finetuning (SFT), trains a base LLM to answer questions and follow instructions by training on a dataset of prompt-response pairs with a language modeling objective. Preference optimization further improves the model's ability to follow conversations and teaches a model to generate responses that align with human preferences by showing the model samples of desirable and undesirable responses (or a ranking of responses). Direct preference optimization (DPO;~\newcite{rafailov2024direct}) is a reward-free preference optimization technique that optimizes directly on the preference data and does not require training a separate a reward model. It is a popular alternative to reward-based methods such as proximal policy optimization (PPO;~\newcite{schulman2017proximal}) because it is less computationally expensive and achieves promising results.

Post-training LLMs in a multilingual setting is an under-explored topic~\cite{ustun2024aya,lai2023okapi,martins2024eurollm}. Previous studies have experimented with monolingual and multilingual instruction tuning of multilingual base LLMs~\cite{shaham2024multilingual,chen2024monolingual}. These studies show that monolingual instruction tuning transfers some instruction-following capability to the other languages in the model but is dependent on the amount of multilingual data that the base LLM was trained on. A few studies have investigated multilingual preference optimization~\cite{lai2023okapi,dang2024rlhf}. \citeauthor{lai2023okapi} generated synthetic preference datasets for 26 languages and performed reward-based preference optimization on BLOOM and Llama 7B models. Their results show that preference optimization offers a slight improvment over SFT. \citeauthor{dang2024rlhf}, however, point out that these preference-optimized models still underperform compared to massively multilingual LLMs that are finetuned only with SFT.   

The scarcity of instruction and preference datasets is a major challenge in post-training LLMs for smaller languages. Previous efforts to assemble finetuning datasets through machine translation, crowd-sourcing, and synthetic data generation include ~\cite{ustun2024aya,lai2023okapi,dang2024rlhf}. Evaluating open-ended responses of chat models in small languages is also a challenge. Previous studies have investigated using LLM-as-a-judge in multilingual settings but these studies focused on standard NLP tasks such as summarization and question-answering~\cite{hada2024large,ahuja2023mega}.

\section{Experimental setup}

We use Poro 34B as the base LLM~\cite{luukkonen2024poro}. Poro is trained on 1T tokens of English, Finnish, and code, with 129B tokens for Finnish. We use the Transformer Reinforcement Learning library (TRL;~\cite{vonwerra2022trl}) for instruction tuning and preference optimization. We finetune all of the model parameters in our experiments.\footnote{We experimented with LoRA finetuning~\cite{hu2021lora}, but our results indicated that full finetuning achieved better performance.} 

We use 32 AMD MI250X GPUs in our experiments. For SFT, we use a micro batch size of 4 and a gradient accumulation step of 1, resulting in a global batch size of 128. We use a learning rate of 2e-5 with a warmup rate of 0.1 and finetune for 3 epochs. For DPO, we use a global batch size of 64, learning rate of 5e-7, warmup rate of 0.1, and train for 5 epochs. 

\section{Datasets}

\paragraph{SFT} We use a curated version OpenAssistant 2 ({OASST2};~\cite{kopf2024openassistant}) containing the top-ranked English conversations. This dataset has 4,692 samples.\footnote{The curated dataset is \url{https://huggingface.co/datasets/sablo/oasst2_curated}. The full dataset is \url{https://huggingface.co/datasets/OpenAssistant/oasst2}.}

We translate OASST2 into Finnish using Poro with few-shot prompting. Poro has been shown to produce higher-quality Finnish translations compared to other open MT systems~\cite{luukkonen2024poro}. For this reason, we did not experiment with translations from other open MT models and focus our efforts on the Poro-translated dataset. We use heuristics to clean up the translations. After post-translation cleaning, our OASST2 Finnish data has 4,399 samples. 

\paragraph{DPO} We use the HelpSteer2 preference dataset~\cite{wang2024helpsteer2}, which consists of publicly-sourced prompts and LLM-generated completions\footnote{We initially chose this dataset because it has a commercially-friendly license. Recently, however,~\citeauthor{lambert2024tulu} pointed out that HelpSteer2 includes ShareGPT prompts which have a questionable legal provenance.}. We use the helpfulness scores included in the dataset to obtain 7,221 preference pairs (chosen and rejected responses). We also translate this dataset into Finnish using Poro. After post-translation cleaning, we end up with 6,037 preference pairs in our Finnish HelpSteer2 dataset.

\section{Evaluation}

We use the Instruction Following (IFEval) benchmark to evaluate instruction-following performance~\cite{zhou2023instruction}. IFEval has 541 prompts where a prompt contains verifiable instructions that can be checked with a deterministic program, circumventing the need of an LLM or human as judge. Examples of instructions include adding keywords to the response, formatting the response in JSON, or responding in a specified language.\footnote{See the IFEval paper for the complete list of instructions and their descriptions.}

We translate the IFEval prompts into Finnish using DeepL~\footnote{\url{https://www.deepl.com/}}. IFEval has 31 prompts that require the response language to be in a language other than English. We exclude these prompts for this work due to Poro being constrained to only English and Finnish. We report the results for the remaining 510 prompts only. IFEval reports strict accuracy and loose accuracy where loose accuracy accepts minor transformations in the responses. For the sake of clarity, we report only the strict accuracy in this work. 

We run evaluations through the LM Evaluation Harness~\cite{eval-harness}. The translated IFEval is available at~\url{https://huggingface.co/datasets/LumiOpen/ifeval_mt}. 

\section{Experiments}

\paragraph{Multilingual SFT}
Finnish instruction data is more difficult to obtain compared to English; therefore, we want to investigate how the amount of Finnish instruction data affects performance. We construct data mixes from the English and Finnish OASST2 datasets such that we start with just the English data and gradually introduce more Finnish samples into the mix starting from 10\% of the Finnish data and then going up to 100\%. We call these data mixes \texttt{en-fi-Xpct} (i.e., the data mix with just the English samples is called \texttt{en-fi-0pct} while the data mix with all the English and Finnish samples is called \texttt{en-fi-100pct}). By default, we do not mask prompts during training (i.e., we incorporate the losses from the prompt and completion tokens). We train SFT models on all the data mixes using the same hyperparameters.

\paragraph{Improving vanilla SFT}
We investigate whether we can improve SFT by adding noise to the word embeddings in the instruction data (NEFTune; ~\newcite{jainneftune}). We also experiment with \textit{prompt masking} where the loss is calculated only on the completion tokens. Our baseline for these experiments is the SFT model trained on the \texttt{en-fi-100pct} data mix.

\paragraph{Multilingual DPO}
We opt to use DPO for preference tuning as it has been found to fare better in IFEval, in addition to being more stable and requiring less compute~\cite{dubey2024llama}. We tuned the $\beta$ parameter of DPO with values $\{0.01, 0.05, 0.1\}$ and found $\beta=0.05$ to be optimal. We experiment with using either the English or Finnish datasets and using both. As our baseline, we use the SFT model trained on the \texttt{en-fi-100pct} data mix.

\section{Results and Discussion}

\begin{table}[t]
\centering
\resizebox{0.5\textwidth}{!}{%
 \begin{tabular}{l | r r b{1.5cm}}
    \textbf{data mix}  & \textbf{EN (\%)} & \textbf{FI (\%)} & \textbf{Resp lang (\%)}\\
    \hline
    \texttt{en-fi-0pct}   & 36.39 & 31.41 & 47.45 \\
    \texttt{en-fi-10pct}  & \textbf{39.97}	& 32.69 & 90.00 \\
    \texttt{en-fi-20pct}   & 37.67	& 28.60 & 93.52 \\
    \texttt{en-fi-40pct}  & 39.20	& 30.90 & \textbf{96.27} \\
    \texttt{en-fi-60pct}  & 39.20	& 32.95 & 94.90 \\
    \texttt{en-fi-80pct}  & 38.56	& 33.84 & \textbf{96.27} \\
    \texttt{en-fi-100pct} & \textbf{39.97}	& \textbf{34.48} & 95.68 \\
 \end{tabular}
 }
 \caption{Instruction-level accuracy on English and Finnish IFEval of the SFT models trained on different data mixes. Response language refers to the proportion of responses classified as Finnish for the Finnish IFEval.}
 \label{tab:ifeval-sft-overall}
\end{table}

\paragraph{Multilingual SFT} 

In Table~\ref{tab:ifeval-sft-overall}, we show the instruction-level accuracy of the SFT models on the English and Finnish IFEval. We also show the proportion of responses to Finnish IFEval that are in Finnish as classified by \texttt{langdetect}\footnote{\url{https://pypi.org/project/langdetect/}}. For English IFEval, the performance is comparable across the data mixes which is expected because the different data mixes contain same number of English samples. For Finnish, the best performance is from the data mix with all the English and Finnish samples (\texttt{en-fi-100pct}). However, compared with the data mix with just 10\% of the Finnish data (around 400 samples), the difference is less than 2 percentage points. If we finetune using only the English data, the resulting model can still follow Finnish instructions but less than half of the responses are in Finnish, which is not desirable.

\begin{figure}[t!]
    \centering
    \includegraphics[width=\linewidth]{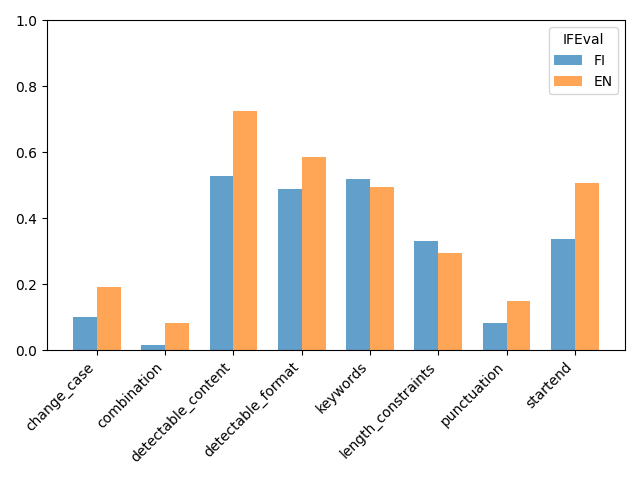}
    \caption{Accuracy by instruction category on English and Finnish IFEval of SFT model trained on the \texttt{en-fi-100pct} data mix.}
    \label{fig:ifeval-eng-fin}
\end{figure}

For the response language, SFT models trained on data mixes containing 20\% and above of the Finnish data have comparable rates of Finnish responses of over 93\%. We reviewed the 22 responses from the \texttt{en-fi-100pct} model that were classified as \textit{not} Finnish. We found that 19 of them are in Finnish but mixed with other languages such as English and German. We also found that the model tends to respond in a mixture of English and Finnish when asked to respond in a specific format, such as JSON or XML. This is likely because JSON tends to be treated as code in the instruction dataset and our translation pipeline did not translate code blocks which sometimes include comments in English. 

Figure~\ref{fig:ifeval-eng-fin} shows the accuracy by instruction category of the \texttt{en-fi-100pct} model. The model struggles most with the combination category---this category includes combined instructions such as giving two responses that are separated by a given separator or repeating the prompt without modification before giving the response proper. The poor performance is probably because the instruction contains multiple steps that must be followed in order to give a correct response. For instance, if the model gives two responses but these responses are numbered instead of separated by an asterisk, the response is considered incorrect. The model also struggles with punctuation instructions such as avoiding the use of commas likely because texts without commas are rare in the dataset.  

Overall, the results indicate that finetuning with a few hundred Finnish instruction samples achieves results close to finetuning with ten times that amount. In terms of instruction types, the model struggles with multi-step instructions and unusual instruction types. Previous work has indicated that carefully curating the instruction dataset is vital to a strong SFT model~\cite{zhou2024lima}. In future, we aim for a smaller but higher-quality data by, for instance, removing highly similar prompts, diversifying tasks, and curating the sources of the samples.

\paragraph{Prompt masking and NEFTune}

In Table~\ref{tab:ifeval-overall-other-experiments} we show the accuracy of the models trained with prompt masking and NEFTune compared to the baseline model.

In our experiments, models trained with NEFTune fail to achieve better scores compared to the plain vanilla SFT baseline. As noted by \citeauthor{jainneftune}, the performance of NEFTune was found to be dataset dependent. One key difference in our study is that we train on a multi-turn dataset. We leave further examination of NEFTune and other noise augmentation techniques for future work. We find that prompt masking does not improve over the baseline. This result is in line with findings from \citeauthor{shi2024} where they show that incorporating the loss from the prompt is beneficial for smaller datasets such as LIMA~\cite{zhou2024lima} with 1,030 examples.

\begin{table}[t!]
\centering
\resizebox{0.5\textwidth}{!}{%
 \begin{tabular}{l | r r b{1.5cm}}
    \textbf{Model}  & \textbf{EN (\%)} & \textbf{FI (\%)} & \textbf{Resp lang (\%)}\\
    \hline
    \textit{baseline} &  \textit{39.97}	& \textit{34.48} & 95.68\\
    prompt masking  & 39.84	& 32.56 & 96.27 \\
    NEFTune            & 38.05	& 32.69 & \textbf{96.47} \\
    DPO-en          & 43.55 & 36.65 & 94.70 \\
    DPO-fi          & 41.76 &  36.01 & 95.49 \\
    DPO-both        & \textbf{44.69}	& \textbf{37.80} & 95.49 \\
 \end{tabular}
 }
 \caption{Instruction-level accuracy on English and Finnish IFEval English for the prompt masking, NEFTune, and DPO experiments. The baseline is an SFT model trained on the \texttt{en-fi-100pct} data mix.}
 \label{tab:ifeval-overall-other-experiments}
\end{table}

\paragraph{Multilingual DPO} 

Table~\ref{tab:ifeval-overall-other-experiments} shows the results from our DPO experiments compared to the SFT model. The model optimized only on English preference data (\texttt{DPO-en}) improved performance on English IFEval by around 3 percentage points and also showed some improvement in Finnish IFEval. This provides further evidence that preference optimization in English benefits other languages in the model~\cite{dang2024rlhf}. The DPO model trained only on Finnish (\texttt{DPO-fi}), on the other hand, showed smaller improvements on both English and Finnish IFEval and, in fact, has slightly lower performance than \texttt{DPO-en} on the Finnish benchmark. The model trained on both languages (\texttt{DPO-both}) achieved the best performance on both benchmarks but compared to \texttt{DPO-en}, the improvements are marginal.  

In terms of the response language, DPO did not improve the Finnish response rates compared to the SFT model. This might be because we optimized the model on monolingual preference pairs (the chosen and rejected responses are in the same language). Improving the response language of multilingual models through preference optimisation is an area we will explore in future work.


\section{Conclusions and Future Work}

In this work we share our findings from post-training Poro 34B in English and Finnish. Due to the scarcity of Finnish post-training datasets we opted to machine-translate instruction and preference datasets using Poro. To evaluate the results of our experiments, we translate IFEval, a widely-used instruction-following evaluation benchmark. We experimented with using different combinations of English and Finnish data in SFT and found that using all available data from both languages gave the best performance overall. Using only 10\% of the Finnish instruction data (around 400 samples), however, still gives competitive performance. We contribute to Finnish LLM development by releasing our datasets, recipes, and model with open licenses at~\url{https://huggingface.co/LumiOpen/Poro-34B-chat-OpenAssistant}.

In future we want to explore different ways of obtaining more Finnish data by, for instance, generating synthetic instruction and preference datasets. We will use these synthetic datasets to further investigate other alignment and finetuning methods. Additionally, we are interested on developing an evaluation benchmark for open-ended conversations in Finnish that takes cultural and linguistic nuances into account. 





\section*{Acknowledgments}

The authors wish to acknowledge CSC – IT Center for Science, Finland, for generous computational resources on the LUMI supercomputer. This project has received funding from the European Union’s Horizon Europe research and innovation programme under Grant agreement No 101070350. The contents of this publication are the sole responsibility of its authors and do not necessarily reflect the opinion of the European Union.

\bibliographystyle{acl_natbib}
\bibliography{nodalida2025}

\end{document}